\documentclass{article}

    \usepackage[preprint]{vendor_neurips18/neurips18}

\usepackage[utf8]{inputenc} 
\usepackage[T1]{fontenc}    
\usepackage{hyperref}       
\usepackage{url}            
\usepackage{booktabs}       
\usepackage{amsfonts}       
\usepackage{nicefrac}       
\usepackage{microtype}      
\usepackage{settings/uniinput} 

\usepackage{pgfplots}

\usetikzlibrary{pgfplots.groupplots}
\pgfplotsset{compat=newest}

\usepackage{soul}
\usepackage{color}

\usepackage{amsmath}

\usepackage{cleveref}

\ifdefined\nohyperref\else\ifdefined\hypersetup
  \definecolor{mydarkblue}{rgb}{0,0.08,0.45}
  \hypersetup{ %
    pdftitle={},
    pdfauthor={},
    pdfkeywords={},
    pdfborder=0 0 0,
    pdfpagemode=UseNone,
    colorlinks=true,
    linkcolor=mydarkblue,
    citecolor=mydarkblue,
    filecolor=mydarkblue,
    urlcolor=mydarkblue,
    pdfview=FitH}

  \ifdefined\isaccepted \else
    \hypersetup{pdfauthor={Anonymous Submission}}
  \fi
\fi\fi
\renewcommand{\cite}[1]{\citep{#1}}
\setcitestyle{numbers,square}

\usepackage{caption,subcaption}  

\newlength\figurewidth
\newlength\figureheight

\makeatletter

\usepackage{catchfilebetweentags}

\usepackage{etoolbox}
\patchcmd{\CatchFBT@Fin@l}{\endlinechar\m@ne}{}
  {}{\typeout{Unsuccessful patch!}}

\usepackage{xifthen}
\let\ORIGinput\input
\renewcommand{\input}[2][]{
	\ifthenelse{\isempty{#1}}{
		\ORIGinput{#2}
	}{
		\ExecuteMetaData[#2]{#1}
	}
}

\newcommand{\thetitle}{Progressive Stochastic Binarization of Deep Networks}

\title{\thetitle}

\author{%
  David Hartmann\\
  Institute of Computer Science\\
  Johannes Gutenberg-University of Mainz\\
  Staudingerweg 9, 55128 Mainz, Germany\\
  \texttt{dahartma@uni-mainz.de} \\
  \And
  Michael Wand\\
  Institute of Computer Science\\
  Johannes Gutenberg-University of Mainz\\
  Staudingerweg 9, 55128 Mainz, Germany\\
  \texttt{mwand@uni-mainz.de} \\
}

\begin{document}

\maketitle

\begin{abstract}
A plethora of recent research has focused on improving the memory footprint and inference speed of deep networks by reducing the complexity of (i) numerical representations (for example, by deterministic or stochastic quantization) and (ii) arithmetic operations (for example, by binarization of weights).\\
We propose a stochastic binarization scheme for deep networks that allows for efficient inference on hardware by restricting itself to additions of small integers and fixed shifts. Unlike previous approaches, the underlying randomized approximation is progressive, thus permitting an adaptive control of the accuracy of each operation at run-time. In a low-precision setting, we match the accuracy of previous binarized approaches. Our representation is unbiased --- it approaches continuous computation with increasing sample size. In a high-precision regime, the computational costs are competitive with previous quantization schemes. Progressive stochastic binarization also permits localized, dynamic accuracy control within a single network, thereby providing a new tool for adaptively focusing computational attention.\\
We evaluate our method on networks of various architectures, already pretrained on ImageNet. With representational costs comparable to previous schemes, we obtain accuracies close to the original floating point implementation. This includes pruned networks, except the known special case of certain types of separated convolutions. By focusing computational attention using progressive sampling, we reduce inference costs on ImageNet further by a factor of up to 33\% (before network pruning).
\end{abstract}

\section{Introduction}
\label{sec:introduction}
The ability to analyze large amounts of training data has been instrumental to progress in machine learning. This applies in particular to deep networks, whose renaissance has been triggered to some extend by the availability of computational resources \cite{LeCun2015,Schmidhuber2015}. Hence, improving computational efficiency is an important endeavor, not only for using current methods on limited hardware such as embedded and mobile systems, but also for further progress in the field \cite{SzeCYE17}.

Current deep networks mostly compute scalar products (weighted sums), i.e., additon and multiplication of real numbers accounts for the largest proportion of computational costs. Of the two named, multiplication is more expensive at a gate-level, as hardware costs grow faster with precision. Nonetheless, floating-point additions also incur substantial costs due to the need for normalization of the mantissa. Thus, hardware vendors have already begun to support specialized low-precision integer and floating-point units. Further specialized hardware has been considered \cite{Ardakani2017IntegralStochastic,Abdelouahab2018CnnFpgaSurvey}. The question of more elementary representations is similarly posed in ``neuromorphic'' computing, which targets non-CMOS hardware \cite{SchumanPPBDRP17} or even machine learning through chemical and biological processes~\cite{Gkoupidenis2017}. 

On the algorithmic side, recent research has explored different approaches for computational cost reduction. Broadly, these can be classified as quantization \cite{GuptaAGN15} (decrease bits per number), binarization \cite{alizadeh2018systematic} (single bit per number), stochastic computing \cite{Gaines1969StochasticComp} (random bit streams), and pruning \cite{HanPTD15} (dropping less important computations) approaches.


\subsection{Our approach}
Our approach combines ideas from stochastic computing, quantization and binarization: 
We represent activations as (small) integers. Weights are binarized stochastically, and multiplied by simple gating. Repeated addition (accumulation) increases the precision when needed. Such a simple stochastic approximation on its own, however, would lead to substantial variance and, thus, large sampling costs. We therefore introduce an importance sampling scheme that performs random choices between adjacent powers of two. While adding only moderate complexity (random 1-bit shifts), this reduces noise dramatically and makes the scheme competitive with continuous representations already at small sample sizes (16 to 64 gated additions already come close to 32-bit floating point accuracy). 

Because this scheme permits trivial local and dynamic accuracy control, we subsequently employ a two-stage algorithm that first computes a rough estimate of accuracy demands (using 8 samples per number) and use the higher accuracy computation only sparsely. This step further reduces sampling costs by factors of 33\% (on \textit{ImageNet} \cite{russakovsky2015imagenet} with a \textit{ResNet50 (v2)} \cite{he2016identity}).

Our representation is particularly suited for the inference stage (forward pass), where the sampled factors (powers of two/shifts) are constants. We therefore evaluate our scheme for inference in common deep networks for image classification, including ResNet \cite{he2016identity}, DenseNet \cite{HuangLMW17}, Inception \cite{SzegedyIVA17}, XCeption \cite{chollet2017xception}, MobileNet \cite{howard2017mobilenets}, and NasNetMobile \cite{zoph2018learning}. Test results approach already at moderate levels of sampling (16 samples) higher precision, and converge to 32-bit floating point calculations for increasing numbers of samples. For the 50-layer ResNet (v2) architecture, for instance, we obtain 94.4\% relative accuracy compared to full-precision calculations at 16 random samples and 98.6\% at 64 samples. Less favorable convergence behavior occurs only in case of deep stacks of successive multiplications without accumulation. In our experiments, this occured only for the MobileNet architecture (due the use of a specific type of separable convolution), which is a known issue for other quantization and binarization schemes and can easily be fixed by a slight architectural modification \cite{sheng2018quantization}. Simple pruning \cite{HanPTD15} of the network, which removes redundant computations, does not seem to affect the efficiency of our stochastic approximation scheme.


In summary, our number representation has the following advantages: For the first time, we are able to quantize pretrained networks without retraining and, at the same time, without significant loss of precision. Similarly, it is the first quantization scheme that allows for dynamic control of precision at run-time. This permits an adaptive sampling algorithm that further reduces costs in practice by a substantial factor. Generally, the method is unbiased and convergent to full precision in the sampling limit, therefore permitting high-precision inference on demand by (still moderately) increased computational costs.

\section{Related Work}
\label{sec:relatedwork}

Binarization techniques, and our technique in particular, are at the center of the following topics, each of which has been studied recently.

\paragraph{Quantization}
By reducing the number of bits per operand, and avoiding expensive floating-point units by utilizing number formats with fixed scaling, computational costs get reduced considerably \cite{Courbariaux2014,GuptaAGN15,Zhou2016}. Direct quantization of full-precision pre-trained models, however, leads to a significant loss of accuracy in deep networks. Thus, these methods require a fine-tuning step.
As a result, most techniques apply quantization during training or as fine-tuning of pre-trained full-precision models.\\
One approach increases the significance of low-precision weights by adaptively rescaling intermediate results globally before and after the weight-multiplication \cite{Courbariaux2014}. Instead of optimising the scaling, it is possible to determine optimal global scaling factors for pretrained models by estimating the statistics of incoming data \cite{2015Darrylarxiv:1511.06393}.\\
A second idea uses an iterative discretize-and-retrain technique. Combining this approach with threshold-based pruning produces networks with a much smaller memory footprint \cite{2015Songarxiv:1510.00149}. Another discretize-and-retrain algorithm translates the weights into representations consisting of only shift-operations that are much easier to implement on hardware \cite{2017Aojunarxiv:1702.03044}.\\
More recent works use the \textit{local reparametrization trick} to include stochastic components into a network without loosing the ability to estimate gradients \cite{kingma2015variational}. One representative of this technique replaces weights by continuous relaxed random variables; stress on weights is increased not being on a specified discrete grid \cite{2016Chrisarxiv:1611.00712}. Closely related is the idea of sampling normal distributed preactivations during training instead of discrete random weights \cite{Peters2018,2017Oranarxiv:1710.07739}.\footnote{
    The sum of i.i.d random variables tends towards a normal distribution due to the central limit theorem.
}\\
Our method is an in-place quantization approach, i.e., additional fine-tuning of the model is not required as all weights transform bijectively into their new represenation. Compared to other quantization methods, our method allows for choosing the quality of the representation at run-time.

\paragraph{Binarization}
The limit case of quantization is binarization, where either weights \cite{RastegariORF16} or even both, weights and activations \cite{2016Matthieuarxiv:1602.02830} are constrained to one of two possible values. This reduces costs dramatically, eliminating multiplications (binary weights) or even reducing all arithmetics to logical operations (full binarization, which has considerable impact on accuracy \cite{alizadeh2018systematic}).\\
The first variants used the \emph{straight-through optimizer} \cite{2013Yoshuaarxiv:1308.3432} to enable gradient descent optimization on discrete variables. Fully binarized network architectures with binarized training processes exist \cite{Matthieu2015BinCon, 2016Matthieuarxiv:1602.02830}, however they tend to show a significant performance gap compared to full-precision networks \cite{zhuang2018rethinking}.\\
Further work concentrates on different number representations. As in quantization, global real-valued scaling factors of binary weights represent numbers with greater variance \cite{RastegariORF16}. Another approach learns number representations with binary coefficients by means of a sum of multiple globally scaled binary weights \cite{2017Xiaofanarxiv:1711.11294v1}. More recently, it has been shown that \emph{ResNets} \cite{he2016identity} are well suited for purely binary weights, if implemented with full-precision shortcuts \cite{2018Zechunarxiv}.\\
We propose a stochastic binarization that closes the performance gap of binary neural networks in most of our experiments. To the best of our knowlegde, other binarization techniques require further hyperparameter tuning or add even more hyperparameters \cite{2018Josepharxiv:1809.10463v1}. Our technique changes the number representations bijectively and thus the same hyperparameters apply. One has to mention that our method reevaluates stochastic weights to even out the randomness, while other methods use a single pass during inference. Our experiments show, however, that only a few accumulations are sufficient to achive similar performance as previous methods. In practice, accumulations can be rolled out and computed in parallel.

\paragraph{Hardware Optimized Architectures}
The application to optimize networks for the use on embedded or mobile systems has led to architectures that allow for a trade off between latency and accuracy \cite{howard2017mobilenets, SandlerHZZC18}. Or, to go into low-level optimizations, discussions on new numerical representations exist \cite{hill2018rethinking}. Ternary quantization\footnote{
    Ternary weights are usally constrained to the values $-1, 0$ or $1$.
} \cite{2016Chenzhuoarxiv:1612.01064}, two-bit quantization
\cite{2017Wenjiaarxiv:1701.00485} and integer only multiplications \cite{2017Benoitarxiv:1712.05877}\footnote{
    The authors use 8-bit weights, 32-bit biases and fixed-point numbers for intermediate results.
} are related interpretations of the restrictions of binarization that also allow for better implementation on hardware.\\
Concrete implementations prove faster inference of networks if implemented directly on hardware \cite{2016Griffinarxiv:1602.04283,Abdelouahab2018CnnFpgaSurvey}. Publications that discuss binary neural networks on FPGAs \cite{2017Yixingarxiv:1702.06392}, improved inference on CPUs using SIMD-Instructions \cite{vanhoucke2011improving} and stochastic binary neural networks for near-sensor computing \cite{2017Vincentarxiv:1706.02344} show the interest in such techniques.\\
Our method uses only shift-operations on small integer numbers. For each such shift-instruction one random bit decides which of two shift-operations to use for accumulation. Although promising for hardware-implementations, it remains to show if hardware architectures have a substantial benefit from our proposed method compared to floating point multiplications.


\paragraph{Alternative Network Design}
A topic strongly related to our work is stochastic computation (SC) \cite{Alaghi2014StochasticComputing,Gaines1969StochasticComp}. Here, numbers are approximated by sequences of binary samples whose average corresponds to the intended number. This idea has recently been applied to hardware implementations of deep networks \cite{Ardakani2017IntegralStochastic,Kim2016DynamicEnergyAccuracyTradeOff,Ren2017SCDCNN}. Similarly, stochastic quantization (see above) has also been proven to be a valuable tool to reduce the negative impact of reduced representational efforts.\\
Closely related to SC is the idea of Sum-Product-Networks (SPN), a familiy of probabilistic graphical models that encode joint distributions of their input random variables \cite{poon2011sum}. SPNs are equivalent to arithmetic circuits, and thus are also promising candidates for computationally efficient networks. They have been used successfully for typical deep learning tasks such as natural language processing \cite{cheng2014language}, image classification \cite{sguerra2016image} and image segmentation \cite{rathke2017locally}.\\
The biggest difference of our work to SC and SPNs is the interpretation of data-streams and intermediate results. SC and SPNs interpret data as random data streams. Our approach, in contrast, uses fixed-point numbers for incoming data and intermediate results, while only weights are random variables. This reduces the variance of intermediate results significantly (see \Cref{sec:experiments}). In contrast to the compelling implementations of SC, we prove the feasability of our approach beyond MNIST. To the best of our knowledge, SPNs have not yet been shown to work feasably on large scale image data, like ImageNet, however, the joint-distribution nature of SPNs and their clear semantics allow for new unseen types of queries \cite{bueff2018tractable,hsu2017online,van2019deep,vergari2016encoding}.

\section{Capacitor Units}
\label{sec:capacitors}


Current deep networks typically consist of linear combinations of activations (or inputs) followed by ReLU non-linearities. For simplicity, we consider a single activation in one specific layer of the network.
\begin{equation}\label{eq:mac}
    y =  \operatorname{relu} \left(  \sum_{i=1}^{d} x_i \cdot w_i \right)
\end{equation}
Here, $x$ denotes the $d$-dimensional activation of the previous layer and $\omega_i$ denotes the corresponding $i$-th weight. We omit the bias term for clarity (w.l.o.g., we can set $x_0\equiv 1$).
Convolutional layers have the same basic computational structure except they reuse weights and limit their support.

In addition, architectures add typically batch normalization to each activation after each layer. W.l.o.g. batch normalization is a fixed affine map
\begin{equation}
    \operatorname{bn}(y) = a \cdot y + b,
\end{equation}
where $a,b$ are constants during inference, and dependent on $x$ during training. This mapping can be ``folded'' into the preceding or following linear layer \cite{Jacob2018}. I.e, it is possible to determine a map $w\mapsto w'$ such that batch normalization can be omitted.\\
Our method replaces multiplication by random shifts. We will discuss in \Cref{sec:experiments:imagenet} that it is crucial for our method to fold successive multiplications, as multiplications of random variables induce much higher variance.


\subsection{Stochastic Binarization with Importance Sampling}\label{sec:capacitors:technique}
As discussed in Section~\ref{sec:introduction}, the most expensive operations in terms of chip area in standard implementations are multiplication and floating-point logic. In order to reduce this cost, we combine previous ideas from the quantization, binarization and stochastic computing literature.

\paragraph{Stochastic multiplication}
First, we observe that a multiplication $w \cdot x$ of a real number $w \in [0,1]$ and an integer $x$ can be estimated stochastically by $B_w \cdot x$ where $B_w \in \{0,1\}$ is a Bernoulli random variable with probability $\operatorname{Pr}(B_w = 1) = w$. By construction the mean equals to the original multiplication,
$$
    E[B_w\cdot x] = w \cdot x.
$$
Thus, we substitute all multiplications $w\cdot x$ after folding,
\begin{equation} \label{eq:bernoulli}
    w \mapsto B_w
\end{equation}
and obtain a statistical approximation of the preactivation \Cref{eq:mac}, the argument of the $\text{relu}$-function.

For the $d$-dimensional scalar product $\sum_{i=1}^d x_i\cdot w_i$, this simple estimator stochastically ignores components of the activation vector $x$, eliminating multiplication at the cost of introducing (substantial) variance.

\paragraph{Value-based importance Sampling}
A drawback of networks consisting of only binary weights is the limited range of possible values. A typical observation is that the probabilities $w$ gather around the borders $0$ and $1$ \cite{darabi2018bnn+}.
This reduces the amount of obtainable information of each gradient descent step, as gradients that point outside the weight-domain $[0,1]$ get discarded (various techniques exist which try to reduce this problem \cite{darabi2018bnn+,2017Xiaofanarxiv:1711.11294v1,yin2018arm}.\\
We propose an unbiased encoding that replaces floating point multiplications completely. Floating point representations consist of three parts: $s \cdot 2^e \cdot m$.
The sign $s\in \{-1,1\}$ and the exponent $e\in \mathbb{N}$ adjust the coarse magnitude, the mantissa $m$ (typically $m\in[0,1]$) fixes the more accurate decimal places. Our stochastic number representation that is well-suited for typical deep-learning computation, replaces the mantissa by a stochastic version that takes one of the values: $1$ or $2$. We even out stochasticity of multiplication with these numbers using accumulators and call the new stochastic multiplication \textit{capacitor-unit}. This design utilizes only low-precision integer addition and does not require multiplication.

In more detail, we therefore split each weight into an exponent $e \in \mathbb{Z}$, a sign $s \in {-1,1}$ and a mantissa $p \in [0,1]$ and reformulate any weight $w$ as
\begin{equation}\label{eq:numbersystem}
    w \mapsto \overline{w} := s\cdot 2^e \cdot  \left(B_p + 1\right),\\
\end{equation}
where
\begin{align}
    s &:= \text{sign}\left( w \right), \\
    e &:= \lfloor \log_2 |w| \rfloor \text{ and }\\
    p &:= \frac{|w|}{2^{e}}-1.
\label{eq:stochasticfloat}
\end{align}
Technically, one would choose stochastically the bitshift $\cdot 2^{e+1}$ with probability $p$ and $\cdot 2^e$ with probibility $1-p$ (See \Cref{fig:numbersystem} (a) and (b)).
By construction, the mean of the representation equals to $w$,
$$
    E[\overline{w}] = s\cdot 2^e \cdot \left(\frac{|w|}{2^e}-1 + 1 \right) = s\cdot |w| = w.
$$

\paragraph{Capacitors}
The binarization scheme from \Cref{eq:numbersystem} has a high variance (See \Cref{sec:capacitors:properties}).\\
In order to reduce the variance, we could draw multiple network samples and average the outcome of every sample. Deep networks are, however, non-linear;\footnote{
    This is an important difference to traditional statistical well-known integration approaches such as Monte-Carlo light-transport in computer graphics \cite{Cook1984}. In this setting linearity allows for averaging of end-results.
} therefore, the statistical estimates obtained from plugging Bernoulli samples into Eq.~\ref{eq:mac} are not consistent for multi-layer networks that include non-linearities. Especially gradients w.r.t. the parameters of a probabilistic variable are hard to estimate \cite{2016Chrisarxiv:1611.00712}. The training process can be augmented to account for this \cite{fu2006gradient}, or to reduce the variance of the gradients \cite{yin2018arm}, but even then, inference can suffer from high variance of intermediate results.\footnote{
    This is intuitively clear: binary samples at the input of a highly nonlinear computational network lead to strongly varying outputs.
}\\
Our solution is as follows: We run statistical averaging layer-wise, averaging over multiple samples \emph{before} applying the non-linearity. We call this step a \textit{capacitor}, in allusion to the analog component. Implementation costs are not impacted, assuming that computations are ordered accordingly and computed in parallel. This is the case for wide networks operating on, for example, lots of pixels, independent samples or activations.

Thus, we adapt \Cref{eq:numbersystem} to
\begin{equation}\label{eq:numbersystem:simulation}
    w \mapsto \overline{w}_n := s\cdot 2^e \cdot  \left(\frac{B_{n,p}}{n} + 1\right),
\end{equation}
where $n$ denotes the sample size, and $B_{n,p}$ the $n$-fold binomial distribution with probability $p$.\\
In practice one would continue to sample from bernoulli distributions, for example by rewriting the multiplication to hardware-optimized operations,
\begin{equation}
    x\cdot \overline{w}_{2^n} = s \cdot \left(\sum_{i=1}^{2^n} x \ll (e + B_p^{(i)})\right) \gg n, \quad B_p^{(i)} \text{ i.i.d}
\end{equation}
and where $\ll k$ denotes the $k$-bit left (multiplicational) shift and $\gg k$ denotes the $k$-bit right (divisional) shift.


We provide further notes on our implementation in the supplementary material.

\subsection{Properties}
\label{sec:capacitors:properties}

\begin{figure*}[t] 
    \setlength\figurewidth{.5\linewidth}
    \setlength\figureheight{.3\linewidth}
    \begin{tikzpicture}
\definecolor{nicecol}{rgb}{0.172549019607843,0.627450980392157,0.172549019607843}

\definecolor{color0}{rgb}{0.12156862745098,0.466666666666667,0.705882352941177}
\definecolor{color1}{rgb}{1,0.498039215686275,0.0549019607843137}
\definecolor{color2}{rgb}{0.172549019607843,0.627450980392157,0.172549019607843}
\definecolor{color3}{rgb}{0.580392156862745,0.403921568627451,0.741176470588235}
\definecolor{color4}{rgb}{0.549019607843137,0.337254901960784,0.294117647058824}
\definecolor{color5}{rgb}{0.737254901960784,0.741176470588235,0.133333333333333}
\definecolor{color6}{rgb}{0.0901960784313725,0.745098039215686,0.811764705882353}

\colorlet{nicecol}{color1}
\colorlet{nicecol2}{color2}
\colorlet{nicecol3}{color0}
\colorlet{nicecol4}{color3}

\begin{groupplot}[
    group style={
        group size=2 by 2,
        xlabels at=edge bottom,
        vertical sep=32pt,
        horizontal sep=10pt
    },
    footnotesize,
    xlabel=$w$,
    tickpos=left,
    ytick align=outside,
    xtick align=outside,
    height=\figureheight,
    width=\figurewidth,
]
\nextgroupplot[ylabel={$e=\lfloor \log_2 |\omega| \rfloor$},title=(a) \textbf{Exponents}]

    \foreach \k in {0, ..., 2} {
        \addplot[nicecol,domain=0:1,variable=\r,samples=2,mark=o,thick,mark size=0.125em] ({2^\k*r+(1-r)*2^(\k+1)},{\k});
        \addplot[nicecol,domain=0:1,variable=\r,samples=2,mark=o,thick,mark size=0.125em] ({-2^\k*r-(1-r)*2^(\k+1)},{\k});
        \addplot [nicecol, mark=*,mark size=0.125em] coordinates {({2^\k},{\k})};
        \addplot [nicecol, mark=*,mark size=0.125em] coordinates {({-2^\k},{\k})};
    }
    \foreach \k in {1, ..., 5} {
        \addplot[nicecol,domain=0:1,variable=\r,samples=2,mark=o,thick,mark size=0.125em] ({2^(-\k)*r+(1-r)*2^(-\k+1)},{-\k});
        \addplot[nicecol,domain=0:1,variable=\r,samples=2,mark=o,thick,mark size=0.125em] ({-2^(-\k)*r-(1-r)*2^(-\k+1)},{-\k});
        \addplot [nicecol, mark=*,mark size=0.125em] coordinates {({2^(-\k)},{-\k})};
        \addplot [nicecol, mark=*,mark size=0.125em] coordinates {({-2^(-\k)},{-\k})};
    }

\nextgroupplot[ylabel={$\text{Var}(\overline{w})$},title=(c) \textbf{Variance},ytick pos=right, ylabel near ticks, yticklabel style={anchor=east,xshift=2.5em}]
    \addplot[nicecol2, domain=-8:8, samples=250,thick]{((-2^(2*floor(log2(abs(x))) + 1) + 3*2^floor(log2(abs(x)))*abs(x) - abs(x)^2))};
    \addplot[gray, domain=-8:8, samples=250,thick,dashed]{x^2/8};

\nextgroupplot[ylabel={$p=\frac{|w|}{2^e}-1$},title=(b) \textbf{Probabilities}]
    \foreach \k in {0, ..., 2} {
        \addplot[nicecol3,domain=0:1,variable=\r,samples=2,mark=o,thick,mark size=0.125em] ({2^\k*r+(1-r)*2^(\k+1)},{1-r});
        \addplot[nicecol3,domain=0:1,variable=\r,samples=2,mark=o,thick,mark size=0.125em] ({-2^\k*r-(1-r)*2^(\k+1)},{1-r});
        \addplot [nicecol3, mark = *,mark size=0.125em] coordinates {({2^\k},{0})};
        \addplot [nicecol3, mark = *,mark size=0.125em] coordinates {({-2^\k},{0})};
    }
    \foreach \k in {1, ..., 5} {
        \addplot[nicecol3,domain=0:1,variable=\r,samples=2,mark=o,thick,mark size=0.125em] ({2^(-\k)*r+(1-r)*2^(-\k+1)},{1-r});
        \addplot[nicecol3,domain=0:1,variable=\r,samples=2,mark=o,thick,mark size=0.125em] ({-2^(-\k)*r-(1-r)*2^(-\k+1)},{1-r});
        \addplot [nicecol3, mark = *,mark size=0.125em] coordinates {({2^(-\k+1)},{0})};
        \addplot [nicecol3, mark = *,mark size=0.125em] coordinates {({-2^(-\k+1)},{0})};
    }

\nextgroupplot[ylabel={$\frac{\text{Var}(\overline{w})}{\left|E[\overline{w}]\right|}$},title=(d) \textbf{relative Error},ytick pos=right, ylabel near ticks, yticklabel style={anchor=east,xshift=2.5em}]
    \addplot[smooth,gray, domain=-8:8, samples=250,thick,dashed]{1/sqrt(8)};
    \addplot[smooth,nicecol4,thick,samples at={-8,-7.95,...,-4.05,-4,-3.95,...,-2.05,-2,-1.975,...,-1.025,-1,-0.975,...,-0.525,-0.5,-0.4975,...,-0.26,-0.25,-0.24,...,-0.13,-0.125}] {sqrt((-2^(2*floor(log2(abs(x))) + 1) + 3*2^floor(log2(abs(x)))*abs(x) - x^2))/abs(x)};
    \addplot[smooth,nicecol4,thick,samples at={0.125, 0.13, ..., 0.24, 0.25, 0.26, ..., 0.4975, 0.5, 0.525, ..., 0.975,1,1.025,..., 1.975, 2, 2.05, ..., 3.95, 4, 4.05, ..., 7.95, 8}] {sqrt((-2^(2*floor(log2(abs(x))) + 1) + 3*2^floor(log2(abs(x)))*abs(x) - abs(x)^2))/abs(x)};

\end{groupplot}
\end{tikzpicture}
    \caption{Figure (a) and (b) show of the values of the exponent of the components of the number system. Figure (b) and (c) show the variance and the relative error. In practice, the values near $0$ are only hypothetical; too many shifts of integers always result in the number $0$.}
    \label{fig:numbersystem}
\end{figure*}
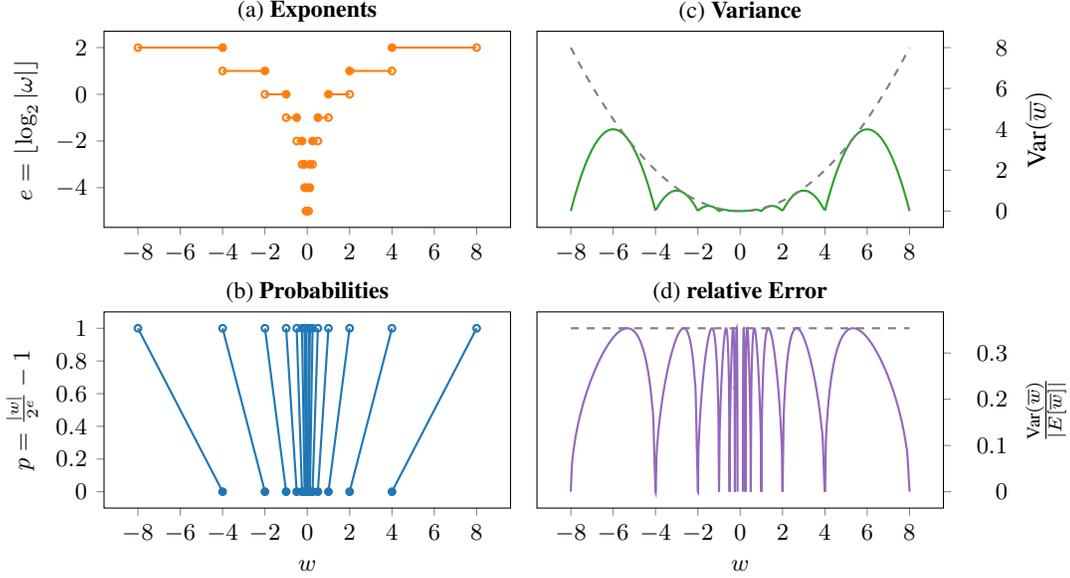

First, we consider the single-sampled case $\overline{w}$.
Even though the mean of $\overline{w}$ equals to $w$, the value varies significantly for large values of $w$ (\Cref{fig:numbersystem} (c)),
\begin{align*}
    Var(\overline{w})
    &≤ \frac{w²}{8}\\
\end{align*}
With multiple samples $n$, the bound decreases antiproportionally,
\begin{equation}
    Var(\overline{w}_n) ≤ \frac{w²}{n\cdot 8}.
\end{equation}

Sampling from this number system leads to high variances between two shifts, i.e.\ it is locally maximal whenever $p = 0.5$. For instance, the representation for $w = 3$ is $(e=1,p=0.5)$. Weights in these regions have high variances (see upper bound above). As our experiments show, however, this is not a problem. We will discuss this in more detail in \Cref{sec:experiments}.

It follows that the \emph{relative variance} of the number system is constant (\Cref{fig:numbersystem} (d)):
\begin{align}
    \frac{σ_{\overline{w}_s}}{|E[\overline{w}_s]|} ≤ \frac{1}{\sqrt{s\cdot 8}}.
\end{align}

\section{Experimental Results}
\label{sec:experiments}
\subsection{Setup}

We implemented the method outlined above in TensorFlow. In all tests, we simulate our method using \Cref{eq:numbersystem:simulation}, thus we sample the corresponding filter directly. We quantize all intermediate results to 16-bit integers ranging from -32 to 32. Absolute run-times are therefore not informative. Instead, we analyse sampling-accuracy trade-offs of our method.

Our experiments are as follows:
First, we study the effect of different globally chosen sampling sizes during training on Cifar-10 (See \Cref{sec:experiments:cifar}), then we study our method as a binarization scheme on pretrained ImageNet (ILSVRC) models (\Cref{sec:experiments:imagenet}). For the latter test case, we additionally examine the effect of probability discretization or naive pruning (\Cref{sec:experiments:pruningandprobability}). Lastly, we implement a naive attention-mechanisms that optimize the overall computational attention, i.e.\ choosing the sample size layerwise or spatially local (\Cref{sec:experiments:attention}).

\begin{figure*}[t] 
    \setlength\figurewidth{\linewidth}
    \setlength\figureheight{.4\linewidth}
    \input{graphics/experiments-eval}
    \begin{minipage}[t]{0.001\textwidth}\end{minipage}\hfill
    \begin{minipage}[t]{0.45\textwidth}
        \caption{Training Cifar-10 using stochastic binarization with different sample sizes. After training, we use the network adaptively with other sample sizes. For reference, we trained the same network using \textit{float32} (the black dashed line) and evaluated it afterwards again using stochastic binarization (the black solid line)}
        \label{fig:smallconv}
    \end{minipage}\hfill
    \begin{minipage}[t]{0.45\textwidth}
        \caption{We evaluate pretrained ImageNet models after binarizing using stochastic binarization with different sample sizes. The dashed lines show the results for their unbinarized floating-point representation.}
         \label{fig:imagenet}
    \end{minipage}
    \label{fig:experiments}
\end{figure*}

\subsection{Cifar-10}
\label{sec:experiments:cifar}
The Cifar-10 dataset consists of 60,000 color images with $32\times 32$ pixels and ten classes of natural objects. Each class consists of 5,000 training and 1,000 test examples \cite{krizhevsky2009learning}.\\
We train a simple eight-layer convolutional network (a stack of eight 3x3 convolutions followed by a batch-normalization and a ReLU-activation) on the Cifar-10 dataset. This low-complexity test permits the study of significant changes of accuracy for this simple task. For the Cifar-10 experiments, we use a normal-distribution based initialization (LeCun initialization). We perform training using Adam optimizer with learning-rate $5e^{-3}$ with exponential decay after each $10$ epochs by a factor of $0.1$, weight decay of $5e^{-4}$, $β_1=0.9$, $β_2=0.999$, $ε=1.0$ for $35$ epochs.

In \Cref{fig:smallconv}, we compare the small convolutional model described above trained with floating point arithmetics to the same model incorporating capacitor units instead. First, we train a full precision network as a baseline (black dashed line in \Cref{fig:smallconv}). As our number representation applies to any weight, we use the pretrained full precision model and evaluate it using our number representation (black solid line) with varying number of samples. In contrast to that, we evaluate the effect of training directly on our progressive stochastic binarization, by training with sample-sizes ranging from $2^0$ to $2^6$ and then evaluating the pretrained weights with other sample sizes of stochastic binarization. Results show that when training using progressive stochastic binarization, the accuracy increased further in contrast to using pretrained models on full precision weights for stochastic quantization in test mode. This is in accordance with other quantization schemes.

\subsection{ImageNet}
\label{sec:experiments:imagenet}
Next we evaluate the binarization on several on ImageNet pretrained architectures in \Cref{fig:imagenet}. ImageNet is
a large natural color image dataset, containing more than 14 million images and over 20,000 image categories. \cite{russakovsky2015imagenet}
We choose the commonly used ILSVRC-Subset. It consists of 1000 object classes and contains approximately 1.2 million training images, 50 thousand validation images and 100 thousand test images.

For evaluation, we use pretrained state-of-the-art architectures,\footnote{The pretrained models are provided by \url{https://github.com/qubvel/classification\_models}} mostly \textit{Inception-} or \textit{ResNet}-Types \cite{he2016identity,SzegedyIVA17}. This constraint is in favour of our implementation, as these architectures provide simple inner structures, namely easily foldable batch-normalizations after convolutional layers. For every model, we include the full-precision performance as a dashed line (see \Cref{fig:imagenet}).\\
In most cases, the binarization scheme already yields half of the accuracy of an unbinarized network with only four samples. With increasing sample size, the results approach floating point calculations.
One of two exceptions is \textit{MobileNet} \cite{howard2017mobilenets}. MobileNet uses separable convolutions to reduce computational costs. But, in contrast to the other mobile-optimized networks \textit{NasNetMobile} \cite{zoph2018learning} and \textit{Xception} \cite{chollet2017xception} that incorporate separable convolutions as well, it seperates depthwise from pixelwise convolutions by relu-nonlinearities, inducing a much higher error due to clipping stochastic results. This is in accordance with previous work that combines MobileNet and quantization \cite{sheng2018quantization}. Also, NasNetMobile and Xception accumulates different intermediate layers, reducing the stochastic errors further.\\
To emphasize the requirement of reducing direct multiplication of stochastic numbers (without convolutions of sufficient kernel size) in long operation chains, we added a modified ResNet model, \textit{Resnet50 modified} with Batch-Normalizations after each shortcut.\footnote{
    The original paper calls this modification “BN after addition“.
}
In this modification, data has to pass multiple Batch-Normalizations, thus, the data-stream that follows the shortcut undergoes multiple batch normalizations that we did not fold into the preceeding layers. Instead, every batch normalizations acts as a seperate multiplication of a stochastic number, additionally to the preceeding stochastic convolution layers. In these shortcut-paths of the graph, the product of successive stochastic approximations increases the variance of the result directly.

\subsection{Weight Pruning and Discretization of Probabilities}
\label{sec:experiments:pruningandprobability}

\begin{figure*}[h]
    \begin{minipage}[b][][b]{0.45\textwidth}
        \small
\begin{tabular}{llrr}
    \toprule
    Experiment & Number & Accuracy \\
               & System & Top-1 [\%] \\
    \midrule
    no modification & \text{float32} & 70.43 \\
                    & \text{psb64} & 69.50 \\
                    & \text{psb32} & 68.56 \\
                    & \text{psb16} & 66.76 \\
                    & \text{psb8 } & 61.86 \\
    \midrule
    pruning 90\% & \text{float32} & 70.34 \\
                 & \text{psb16} & 66.50 \\
    \midrule
    pruning 99\% & \text{float32} & 45.88 \\
                 & \text{psb16} & 39.10 \\
    \midrule
    6-bit probs & \text{psb16}  & 66.64 \\
    4-bit probs & \text{psb16}  & 66.80 \\
    3-bit probs & \text{psb16}  & 66.23 \\
    2-bit probs & \text{psb16}  & 50.29 \\
    1-bit probs & \text{psb16}  & 19.09 \\
    \midrule
    \textbf{attention} & \textbf{psb8/16}  & \textbf{65.74} \\ 
                       & \textbf{psb16/32} & \textbf{68.44} \\ 
    \midrule
    \textbf{combined} & \textbf{psb8/16}  & \textbf{66.15}\\ 
                      & \textbf{psb16/32}  & \textbf{68.29}\\
    \bottomrule
\end{tabular}

        \medskip
        \captionof{table}{
            Classification error (\%) for the same (\texttt{float32}-)pretrained ResNet50 (v2) using the following modifications; plain inference, inference with pruning, inference with probability-discretization, inference using our attention mechanism and inference of a combination of the named techniques. We used two number systems: \texttt{float32} and ours (\texttt{psb16}, 16-fold sampling) for inference.
        }
        \label{tab:modifications}
    \end{minipage}\hfill%
    \begin{minipage}[b][][b]{0.45\textwidth}
        \begin{center}
            \begin{minipage}{0.6\textwidth}
                (a)\hfill(b)
            \end{minipage}
            \vspace{-0.1cm}
        \end{center}
        \includegraphics[scale=0.393]{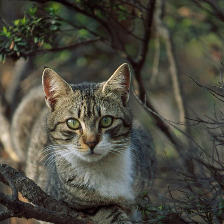}
        \includegraphics[scale=0.393]{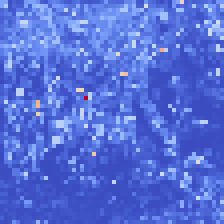}\vspace{0.1em}
        \includegraphics[scale=0.258]{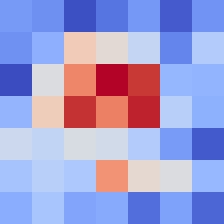}
        \includegraphics[scale=0.258]{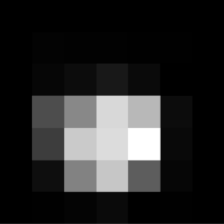}
        \includegraphics[scale=0.258]{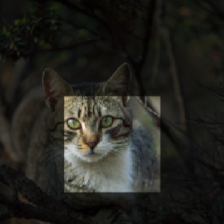}
        \begin{center}
            \vspace{-0.12cm}
            \begin{minipage}{0.73\textwidth}
                (c)\hfill(d)\hfill(e)
            \end{minipage}
        \end{center}
        \captionof{figure}{
            For a given image (a), we show the mean pixelwise approximation errors of 2 samples of our number system compared to \texttt{float32} calculations in the first (b) and the last convolutional layer, $L$, (c), in the ResNet50 (v2) network.\\
            We estimate the importance of the region by computing the pixelwise entropy in the layer $L$, (d), and its resulting mask, (e), for 8 samples. Blue or dark areas represent smallest values in the layer, red or light areas represent the biggest values in the layer.
        }
        \label{fig:error}
    \end{minipage}

\end{figure*}

Next, we evaluate typical graph modifications to allow for even more computational or memory-wise efficiency on hardware.
For these tests, we focus on a single respresentative network pretrained on ImageNet. We chose the ResNet50 (v2) model as a commonly used respresentative for image-recognition tasks. For simplicity, we evaluate only float32, and, our discretization with 16 samples. \Cref{tab:modifications}\\
First, we apply a straight-forward magnitude-based threshold pruning method \cite{HanPTD15} to reduce 90\% and 99\% of all weights close to zero.
In \Cref{tab:modifications}, we observe that pruning of 90\% of weights does affect stochastic computation reasonably, however, for over-pruning (without retraining) of 99\% of weights, disadventages are more noticable for stochastic computation.\\
Next, we reduce the memory footprint of the network. As stated in \Cref{sec:capacitors}, using a high-precision probability-value of the number system, $p$, does not impact computational performance under the assumption that random numbers are generated efficiently. The reason for that is, that our method uses each probability value to generate one single bit to choose between one of two shifts. A drawback is, however, that high-precision probability-values do impact the memory-footprint of the network. Thus, we evaluate a strict quantization of probabilities in \Cref{tab:modifications}. We reduce the number of bits for each probability to $6$, $4$, $3$, $2$ bits and discrete quantization (i.e.\ $1$ bits), where the quantization of probabilities is regular, including the boundary $p=0$ and excluding the boundary $p=1$ (the right boundary would result in a higher exponent). The results indicate that for the weight discretization, the accuracy does not change much in a stochastic setting, but drops significantly for the discrete case ($1$-bit).
As we use $16$-bits fixed point numbers for all intermediate results, we conclude that 4-bit exponents and 4-bit probabilities are sufficient for the use of our progressive stochastic binarization scheme on typical image recognition tasks.

It follows that our method can be reformulated to a deterministic version for tasks that require only a limited upper precision. For instance, for a $4$-bit quantization of probabilities, we cannot gain any accuracy of the number system for more than $16$ samples. Thus, instead of sampling the probability $p=3/16$, we could use deterministically the smaller shift in $3$ of $16$ cases. This deterministic version, however, does not allow for a dynamic control of higher precision calculations.

\subsection{Computational Attention}
\label{sec:experiments:attention}
In this section, we take advantage of the adaptive sampling property of our method. As above, we use a ResNet50 (v2) model, pretrained on ImageNet. First, we discuss if computational effort benefits from a layer-wise adaption of sample accuracy. Then, we evaluate a simple spatial sampling adaption method.\\
One way of doing so is to adapt the sample size of each layer separately. In \Cref{fig:error}, we analyse the approximation error, $|\frac{x_\texttt{psb2}-x_\texttt{float32}}{x_\texttt{float32}}|$, of the network using our method (\texttt{psb2}) compared to floating point calculations. We use our number system in a low-precision regime of only $2$ samples per multiplication. For a given image (\Cref{fig:error}a), we estimate the mean pixelwise approximation error using 100 samples of inference up to the first convolutional layer (\Cref{fig:error}b) and the last convolutional layer (\Cref{fig:error}c).\\
We evaluated if the distribution of errors changes with the layers (i.e. larger errors in lower layers or vice versa), however, we did not notice any such characteristic.

The approximation error seems to follow local features, at least in the last layer. In the following, we inspect the feasability of spatial adaption of computation. Thus, we optimize the overall computational attention by evaluating the network in a high-precision mode only on the regions of high activation in the image. We obtain those regions by using the same network in a low-precision mode on the full image (8 samples per computation). In more detail, we evaluate the pixelwise entropy of the image in the last convolutional layer (\Cref{fig:error}d): We estimate the entropy $h_{xy}$ by
$$
    h_{xy} := \sum_{c} - \text{softmax}(a_{xyc})\cdot \log(\text{softmax}(a_{xyc})),
$$
where $a_{xyc}$ denotes the activation of the last layer in the pixel $(x,y)$ and the channel $c$. In conclusion, regions of lowest pixelwise entropy have high magnitudes in only a few channels, regions of highest pixelwise entropy have equal activations in all channels. Thus, we have to refine high-entropy regions as those have higher probability of returning the wrong classification. For our attention mechanism, we use a hard threshold at the mean entropy in the image to estimate the interesting regions (\cref{fig:error}e). For the ImageNet testset this results in a ratio of about 35\% of interesting regions of higher entropy and 65\% of regions of lower entropy.\\
We use these two types of regions to seperate the activations spatially (on all layers), to adapt computational attention of each filter individually (\Cref{tab:modifications}, attention modification). In the first experiment, we use 8 samples for regions outside computational interest and 16 samples on these regions (psb8/16). In total, we reduce computational effort by a factor of 33\% compared to psb16. We expect this factor to increase significantly for datasets with higher image dimensions and thus, smaller regions of interest, like the MSCOCO-Dataset. The second experiment (psb16/32) increases computational effort by a factor of 33\% compared to psb16, but results in performance close to psb32, a sampling mode of twice as many samples as psb16.\\
We reduce computational costs further by applying the previous techniques, probability discretization (to 4-bit probabilities) and pruning (90\% threshold based pruning of weights) without losing significant amounts of precision (\Cref{tab:modifications}), resulting in a total of 68.29 \% top-1 accuracy on the ImageNet test dataset.

\section{Conclusion}
\label{sec:conclusion}
We have introduced progressive stochastic binarization (psb), an unbiased stochastic binarization scheme for deep networks that replaces all floating-point multiplications. Our method converts network weights into stochastic shift operations. We even out stochasticity using layer-wise sampling. Progressive stochastic binarization therefore allows for efficient inference on hardware by restricting itself to additions of small integers and fixed shifts. The individual operations employed can be implemented on CMOS hardware at reduced costs (a full evaluation on actual FPGAs or ASICS, however, is still subject to future work), and the key operation of gating of continuous information might also be a useful primitive for unconventional non-CMOS computing processes such as neuromorphic circuits. We applied the method for inference of pre-trained image classification networks, and also during training of newly initialized models. Performance wise we match the accuracy of previous binarized approaches in a low-precision setting, and in a high-precision regime, our method is accuracy-wise competitive with previous quantization schemes. The major limitation is that although our scheme applies successfully to a wide range of ResNet architectures, we found that it cannot be applied to specific types of separable convolutions. We examined the best practices for architectures by modifying a previously well-working ResNet architecture. Nonetheless, the method also permits localized, dynamic accuracy control within a single network, providing a new tool for adaptively focusing computational attention; only few quantization schemes provide this feature. We use that feature to direct computational costs adaptively using the network itself as an attention proposal mechanism for better classification results.

\bibliography{bibliography}

\bibliographystyle{vendor_icml19/icml2019}

\clearpage

\title{Supplementary Material for: \\ \thetitle}
\maketitle
\setcounter{section}{0}

\section{Notes on our Implementation}

\textbf{Forward pass:} We obtain the forward pass by substituting the matrix products that linearly combine activations using the weights by capacitor units, as described in \Cref{sec:capacitors:technique} of the paper.\\
Our implementation simulates the method, thus it performs all computations using \texttt{float32} precision. In order to account for quantization, we quantize to 16-bit fixed-point numbers, ranging from $-32$ to $32$.\footnote{
We provide a tensorflow-tool that first folds all batchnorm-convolution pairs, then converts all remaining convolutions and batch-normalization layers to our stochastic number representation, and quantizes all bias terms and intermediate results to fixed-point numbers.
}
On all of our experiments, 16-bits were enough, however, using less bits was sufficient for some architectures as the values of the activations did not reach the boundaries $±32$.

\textbf{Backward pass:} 
Our method supports an optional training step to improve performance for networks that have not been adapted to stochastic quantization noise. We currently perform training on standard hardware (GPUs or CPU).
In the backward pass we just optimize continuous weights and convert them into the exponent-probability representation after each step of gradient descent (regardless of the specific method such as Adam, Momentum or SGD). We compute gradients without any modification; as we introduce a new number representation we calculate the gradients as if no modification was made to the weights.

\textbf{Efficient simulation of the forward pass:} 
A technical problem is that GPU hardware does not benefit from our optimizations; to the contrary, despite already running \texttt{float32}-computations, additional costs incur for quantization and format conversion. In addition, if not parallelized, repeated sampling slows down the process linearly, without achieving any speed benefits for each individual pass. We therefore augment the forward pass to directly sample from Binomial distributions. Rather than accumulating $n$ Bernoulli experiments with probability $p$, we sample from
\begin{equation}
    B_p(k) = \binom{n}{k}p^k(1-p)^{n-k}
\label{eq:binomial}
\end{equation}
using the Gumbel-Max-Trick \cite{gumbel1954statistical};
\begin{align}
    B_p(i) &= \arg \max_k \left[ \log \left(\binom{n}{k}p^k(1-p)^{n-k}\right) - \log\left(- \log U_k\right),\right]\\
\end{align}
where $U_k$ denotes i.i.d. uniform variables $U_k \sim U(0,1)$. We reformulate this to be numerically stable by using $\log$-rules,
\begin{align}
    B_p(i) &= \arg \max_k \left[ \log \binom{n}{k}+k\cdot \log(p) + (n-k)\cdot \log(1-p) - \log\left(- \log U_k\right),\right].
\end{align}

\subsection{Hardware Implementation Perspectives}

\begin{table}[t]
    \centering
    \small
    \begin{tabular}{lrlr}
    \textbf{operation} & \textbf{chip area} & \textbf{chip area, relative} & \textbf{energy} \\
                       &  [$\mu m²$] & to \texttt{fp32 mul} & [$pJ$] \\
    \hline
    \texttt{int8 add}  & 36    & ~0.005 & 0.03 \\
    \texttt{int16 add} & 67    & ~0.01 & 0.06 \\
    \texttt{int32 add} & 137   & ~0.02 & 0.10 \\
    \texttt{int8 mul}  & 282   & ~0.04 & 0.20 \\
    \texttt{int32 mul} & 3,495 & ~0.45 & 1.10 \\
    \hline
    \texttt{fp16 add}  & 1,360 & ~0.18 & 0.40 \\
    \texttt{fp16 mul}  & 1,640 & ~0.21 & 1.10 \\
    \texttt{fp32 add}  & 4,184 & ~0.54 & 0.90 \\
    \texttt{fp32 mul}  & 7,700 & 1 & 3.70 \\
    \hline
\end{tabular}

    \caption{Hardware costs for common arithmetic units (45nm-process, \cite{Dally2016,Horowitz2014}).}
    \label{tab:costs}
\end{table}
As the computational structure of deep network training and inference is highly data-parallel, a reduction in chip area and power consumption per operation directly translates to substantial performance advantages (see \Cref{tab:costs}).\\
The representation outlined in the main part of the paper is motivated by gaining potential for a more efficient implementation on custom CMOS hardware.

\begin{figure*}[h] 
    \centering
    \includegraphics[scale=0.15]{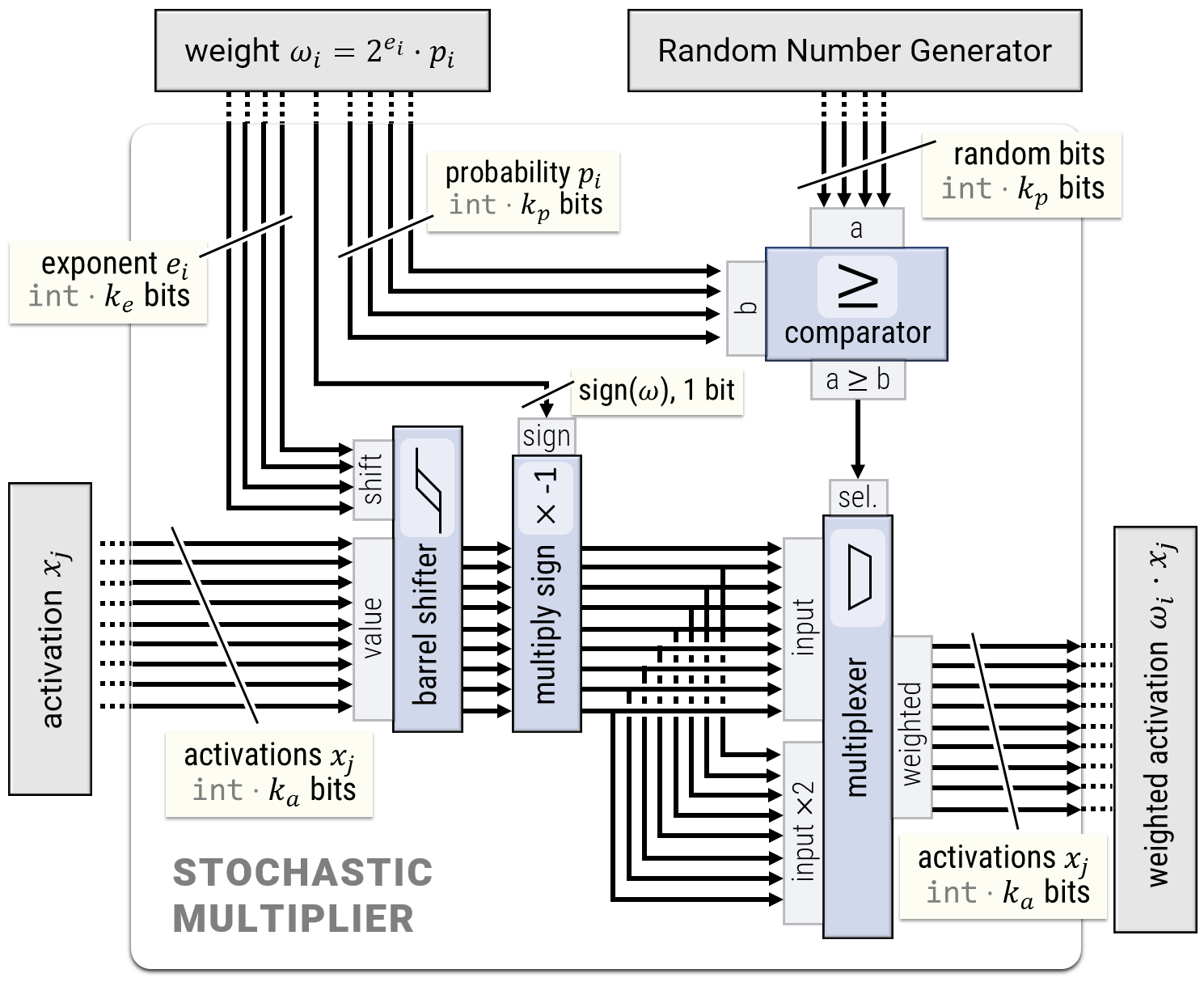}
    \\
    \begin{minipage}[t]{1.00\textwidth}
        \caption{
             Schematic circuit diagram of a stochastic multiplier.\\
             It performs multiplication by randomly selecting between two shifted versions of the activation; all shifts are by constants. One single random bit with probability $p$ chooses wich of those two shifts ($e$ or $e+1$) is used.
             The accuracy of intermediate integer representation restricts the magnitude of the exponents, the precision of the probability values is restricted by the memory requirements and accuracy requirements of the network.
        }
        \label{fig:StochasticMultiplier:stochmult}
    \end{minipage}
\end{figure*}

\textbf{Stochastic multiplier:} In \Cref{sec:experiments}, we quantized probabilities using $k_p$ integer numbers. Sampling of random bits with (quantized) probability $p\in [0,1]$ requires a $k_p$-bit comparator, which corresponds to an accordingly sized integer subtraction unit. This subunit returns one single random bit. As no other operation depends on the concrete value of $p$, the number of bits $k_p$ is only bound by memory restrictions. In \Cref{sec:experiments} of the main paper, we have discussed the implications on performance if reducing this number dramatically.\\
Handling exponents (stored as $k_e$-bit integers) for every weight requires a barrel-shifter (i.e., the shift is selected at run-time by turning on or off shifts by powers of two by control-bits); the additional shift by two is obtained by simple wiring. Finally, a multiplexing gate, controlled by the randomly sampled bit, determines the outcome. In \Cref{sec:experiments}, we observe that 4-bit exponents and 4-bit probabilities are sufficient in all our experiments.

\textbf{Activations and Accumulation:} We store all activations and perform all accumulations in integer format (fixed point). As addition units are cheap compared to multiplications, this appears to be a better trade-off than reintroducing adpative shifts of a full floating-point scheme. Our Experiments show that a 16-bit integer representation is sufficient to avoid signifigant accuracy penalties. \Cref{fig:StochasticMultiplier:stochmult} shows a rough data-flow diagram of a full capacitor unit.

\textbf{ReLU, Batch-Norm, and Pooling:} ReLU units are easy to realize in hardware (using a gate that depends on the sign bit). Batch-norm requires a more sophisticated approach; during training, one could quantize factors to powers of two. When using pretrained networks, this is not possible (in particular when short-cut connections void scaling invariance). Thus, the better alternative is to ``fold'' the mapping into the linear weights as stated in \Cref{sec:capacitors}. Note that we can fold the batch-norm into the weights before the calculation of the encoding, i.e.\ we transform $e$ and $p$ accordingly. Similarly, non-power-of-two scaling factors in average-pooling layers can be folded for pretrained networks as well, if needed, or implemented as multiplication with a global stochastic number (both variations did work in our experiments). Max-pooling needs additional comparator units and gating.

\textbf{Random number generation:} Stochastic computation requires random numbers that are sufficiently independent and evenly distributed. Simple linear feedback shift registers are sufficient for this purpose; however, they add some overhead (numbers are generated by recurrent fixed shifts, additions, and XOR with constants) \cite{marsaglia2003random}. The current implementation of tensorflow uses \textit{XORWOW} as the standard random number generator on graphics cards and \textit{MT19937\_64} for CPU-operations. We tested both and did not recognize any differences in the resulting performance.

\textbf{Classification Layer:} For inference, we can ignore the softmax-layer, as we are only interested in the maximum value of the classification. A possible average pooling in the last layer can be ignored w.l.o.g., as the normalization can be trivially folded to the preceeding layer.\\
Training, however, may need a softmax-layer depending on the loss function. We did not cover this in the scope of this work; for simplicity, we did not replace these layers in our implementation.

\textbf{Systolic Arrays:} Modern Computational Units like GPUs and TPUs include systolic arrays for faster matrix-matrix multiplication \cite{kung1982systolic}. It appears plausible that our method can be included in various ways directly into systolic arrays, as the accumulation step of independent samples and matrix-multiplication are interchangeable due to linearity.\\
Thus, one could directly replace the innermost floating point multiplication of systolic arrays with a capacitor unit. For more flexibility sampling-wise, however, it is beneficial to replace the multiplication with a single stochastic multiplier without accumulation (\Cref{fig:StochasticMultiplier:stochmult}). For multiple samples, one could either sum any number of results of the single-sampled systolic array, or concatenate any number of same weight tensors as the input.

\textbf{Discussion:} The proposal leaves out many details required for an actual hardware design, such as allocating the right ratio of accumulation and multiplication units, organizing the accumulation loop, pipelining, and memory management (storage for intermediate results, avoiding energy costs for non-local memory, and many more). Our paper focuses on the main representation and its accuracy properties, in particular in practical experiments. Weighting the costs against alternative quantization schemes might lead to trade-offs that can only be assessed in a full hardware implementation. The idea of stochastic accumulation is, however, of conceptual interest beyond digital CMOS circuits.\\
Neuromorphic computing with (partially) analog elements could similarly benefit from it, and accumulation could be performed by capacitors (electrical or otherwise), and sources of random noise might be available without digital computation.

\end{document}